\documentclass{article}

\usepackage{arxiv}

\usepackage{amsmath,amssymb,amsthm}
\usepackage{nicefrac}
\usepackage{xspace}
\usepackage{epsfig}
\usepackage[shortlabels]{enumitem} 

\usepackage{color}
\usepackage{url}
\usepackage{booktabs,multirow}
\usepackage{tikz}
\usepackage{verbatim}
\usepackage{csquotes}
\usepackage{wasysym}
\usetikzlibrary{arrows,shapes}

\usepackage{pifont}

\usepackage{algorithm}
\usepackage[noend]{algpseudocode}

\algrenewcommand\alglinenumber[1]{\scriptsize #1:}

\usepackage{tikz}
\usetikzlibrary{calc, positioning, arrows.meta}

\newcommand{\pflipcustomer}{p_a}
\newcommand{\pflipvehicle}{p_v}
\newcommand{\pswap}{p_\text{swap}}
\newcommand{\nswap}{n_\text{swap}}

\newcommand{\CA}{C}
\newcommand{\CM}{M}
\newcommand{\CD}{D}
\newcommand{\CDnewavailable}{\CD^{\text{new}}}
\newcommand{\Cavailable}{\CA^{\leq t}}
\newcommand{\CDavailable}{\CD^{\leq t}}
\newcommand{\tours}{T}
\newcommand{\dtours}{T^{\leq t}}
\newcommand{\eralength}{\Delta}
\newcommand{\neras}{n_t} 
\newcommand{\nvehicles}{n_v}
\newcommand{\population}{P}
\newcommand{\popsize}{\mu}

\newcommand{\sdom}{\prec}

\title{Dynamic Bi-Objective Routing  of Multiple Vehicles}

\author{
   Jakob Bossek \\
  Optimisation and Logistics\\
  The University of Adelaide\\
  Adelaide, Australia \\
  \texttt{jakob.bossek@adelaide.edu.au} \\
   \And
  Christian Grimme \\
  Information Systems and Statistics \\
  University of M{\"u}nster \\
  M{\"u}nster, Germany \\
  \texttt{grimme@wi.uni-muenster.de}
  \And
  Heike Trautmann \\
  Information Systems and Statistics \\
  University of M{\"u}nster \\
  M{\"u}nster, Germany \\
  \texttt{trautmann@wi.uni-muenster.de}
}

\begin{document}
\maketitle

\begin{abstract}
In practice, e.g. in delivery and service scenarios, Vehicle-Routing-Problems (VRPs) often imply repeated decision making on dynamic customer requests. As in classical VRPs, tours have to be planned short while the number of serviced customers has to be maximized at the same time resulting in a multi-objective problem. Beyond that, however, dynamic requests lead to the need for re-planning of not yet realized tour parts, while already realized tour parts are irreversible. In this paper we study this type of bi-objective dynamic VRP including sequential decision making and concurrent realization of decisions. 
We adopt a recently proposed Dynamic Evolutionary Multi-Objective Algorithm (DEMOA) for a related VRP problem and extend it to the more realistic (here considered) scenario of multiple vehicles. We empirically show that our DEMOA is competitive with a multi-vehicle offline and clairvoyant variant of the proposed DEMOA as well as with the dynamic single-vehicle approach proposed earlier.
\end{abstract}


\section{Introduction}
\label{sec:introduction}

Routing of multiple vehicles is an important and difficult problem with applications in the logistic domain~\cite{schmid2013rich}, especially in the area of customer servicing~\cite{Flatberg2007}. In postal services, after-sales services, and in business to business delivery or pick up services one or more vehicles have to be efficiently routed towards customers.
If customers can request services over time, the problem becomes dynamic: besides a set of fixed customers, new requests can appear at any point in time. Of course, it is desirable that as many customers as possible are serviced while the tour of any vehicle is kept short. However, it is usually infeasible (due to human resources, labor regulations, or other constraints) to service all customer requests. And clearly, the less customers are left unserviced, the longer the tours become. Thus, the problem is inherently multi-objective. Any efficient solution (smallest maximum tour across all vehicles) is a compromise between the desire to service as many customers as possible (e.g. maximize revenue) and the necessity to keep vehicle routes short (minimize costs). At the same time, the dynamic appearance of new customer requests may significantly change the scenario over and over again: new but ignored requests negatively contribute to the objective of visiting as many customers as possible, while the inclusion of new customers (usually) increases tour length and thus changes the compromises on which a selection of a route was originally made by a decision maker (DM).

This dynamic problem has been studied by Bossek et al.~\cite{BGMRT2019BiObjective} for the special case of a single vehicle which answers all requests and travels (in an open tour) from a start to an end depot. The authors devised a dynamic evolutionary approach based on an interactive algorithmic framework that incorporated an evolutionary multi-objective optimization algorithm (EMOA) applied in eras and repeated decision making. However, the applied EMOA is rather unrealistically based on the assumption, that only one vehicle is available. 

In this work, we will reuse the framework proposed by Bossek et al.~\cite{BGMRT2019BiObjective} but replace the internal EMOA~\cite{BGMRT2018} by an adapted algorithm that is capable of considering multiple vehicles. The inclusion of multiple vehicles changes the problem (and thus the algorithm) considerably: Instead of a single tour (single open TSP),
as many tours as considered vehicles have to be optimized simultaneously. This implies changes in problem encoding, in information transfer between generations, and in variation operators.

At the same time, the number of vehicles is explicitly not considered as additional objective. To keep the scenario realistic, the number of vehicles can neither be changed during the optimization process nor in each era. A dynamic change in the number of used vehicles during the process would require most flexible (and thus costly) human resources and is therefore usually infeasible for a company.\footnote{Visits at single or few customers (including direct travel from and return to a depot) would immediately contribute high costs to the total tour costs.
} Additionally, a third objective would turn the originally bi-objective problem in a more complex-to-handle decision scenario for the DM.

The goal and contribution of this work is twofold:
\begin{enumerate} 
    \item The dynamic multi-objective vehicle routing problem (MO-VRP) and the dynamic solution approach are extended towards a more realistic scenario by including multiple vehicles. We introduce a significantly changed algorithm within the interactive framework proposed in~\cite{BGMRT2019BiObjective}.
    \item We analyze the benefit of multiple vehicles in dynamic vehicle routing and compare our approach to an (extended) version of an a-posteriori evolutionary solution approach for this problem~\cite{GMT+2015}. This approach unrealistically knows of all service requests in advance (clairvoyant) and thus needs no dynamic decision making during optimization. 
    Further, we compare the multi-vehicle approach to the dynamic approach by Bossek et al.~\cite{BGMRT2019BiObjective} and investigate the activities of vehicles. The individual activities of vehicles provide information on whether each vehicle contributes to the solution or whether some vehicles stay idle. This evaluation can eventually justify our decision to not include the number of vehicles as third objective.
\end{enumerate}

The work is structured as follows: the next section briefly reflects the related work, while Section~\ref{sec:probform} formally introduces the dynamic multi-objective problem as described by Bossek et al.~\cite{BGMRT2019BiObjective}. Section~\ref{sec:demoa} then details the algorithmic extensions. The experimental setup as well as empirical results are described and discussed in Sections~\ref{sec:experiments} and \ref{sec:experimental_results}. Section~\ref{sec:conclusion} finally concludes the work and highlights perspectives for future research.

\section{Related work}
\label{sec:related}
As the traveling salesperson problem (TSP) is a major sub-problem of the here considered dynamic and multi-objective vehicle routing problem, this paper is naturally related to work on special TSPs, where not all customer locations (or cities) have to be visited. In research, these problems are sometimes referred to as orienteering problems~\cite{GLV87}, selective TSP~\cite{GLS98a,LM90}, or as TSP with profits~\cite{FDG05}. However, most of these problems are discussed as single-objective problems~\cite{DMV95}, although some early work already recognized the (at least) bi-objective character of these problems~\cite{KG88}.
Only later work started to solve the orienteering problem in a bi-objective way using an $\epsilon$-constraint approach~\cite{BGP09} or approximation schemes~\cite{FS13} that produce Pareto-$\varepsilon$-approximations of the efficient solution set. While the both before mentioned approaches are based on repeated single-objective optimization, some authors \cite{Ombuki2006,JGL08,tan2006,kang2018enhanced,wang2018bi} explicitly solve the bi-objective variants of the orienteering problem using an evolutionary algorithm, however, excluding service requests over time or considering the problem as a-posteriori (non-dynamic). Many of these approaches~\cite{Ombuki2006,tan2006,kang2018enhanced} introduce the number of vehicles as an objective to be minimized while simultaneously minimizing the tour length.

A related a-posteriori variant of the here considered dynamic problem is described by Grimme et al.~\cite{GMT+2015}, who propose an NSGA-II-based EMOA. This work has been extended later on by the integration of local search mechanisms~\cite{MGB2015} and the analysis of local search effects~\cite{BGMRT2018}. These works only allow one vehicle (like also described in \cite{GLV87,VSV11}) but include the number of visited customers (revenue) as second objective besides tour length (costs). 

While considering only one vehicle seems to be unrealistic, the inclusion of the number of vehicles as objective is only feasible in the a-posteriori and non-dynamic case. When problem instances change constantly due to customer requests (related examples from logistics and other domains can be found here: \cite{GFK17,pinedo2012scheduling,MPGL06}) decision making is also a repetitive process over time. However, over time, decisions are constantly renewed building on past decisions which of course cannot be changed. In vehicle routing, one or more vehicles start at a depot and travel initially decided tours. Later on, new decisions have to take into account the current location as well as newly received or not yet serviced customer requests~\cite{BGMRT2019BiObjective}. Rewinding of previous decisions (i.e. visits of customers) is impossible. As such, the initial decision for a fixed number of vehicles could only be changed by sending vehicles home or activating additional ones. This however, causes additional traveling costs and contradicts (in the real world) human resources' availability or labor regulations. Thus, it is most realistic not to consider the number of vehicles as additional objective in the dynamic case.

In general, dynamic vehicle routing is usually addressed by designing online decision rules, see e.g.~\cite{Pillac2013,Meisel2011}. According to Braekers et al.~\cite{Braekers2016} only little work is available on dynamic multi-objective problems. In their survey they mention authors who consider dynamics in service time windows and changing structures of the network~\cite{Wen2010,Lorini2011,Khouadjia2012,Hong2012,Barkaoui:2013}.

\section{Problem notation}
\label{sec:probform}

The here considered dynamic multi-objective VRP can be denoted as follows: we consider a set of customer locations, which can be partitioned into three disjoint sets, $\CA=\CM\cup\CD\cup\{N-1, N\}$. The subset $\CM$ contains all \underline{m}andatory customers locations that are initially known and have to be visited, while the subset $\CD$ contains all \underline{d}ynamic customers that appear over time and are not known to the algorithm beforehand. The third subset $\{N-1,N\}$  denotes the locations of the start and the end depot. Note that we consider the more general case here, in which start and end depot can be different. The more common special case of start and end depot being at the same location is of course included.


We consider two objectives in a minimization problem. The first objective aims for the minimization of the maximum tour length for all vehicles. Let $\nvehicles$ be the number of vehicles and $x$ a solution to the problem, then we denote the tour length for each vehicle $i\in\{1\dots,\nvehicles\}$ as $L_i(x)$ and determine $f_1(x)=\max_i L_i(x)$. By using the maximum tour length as objective, we expect a balanced usage of vehicles in any solution.
The second objective $f_2$ minimizes the number of unserved dynamic customers\footnote{Note that we consider the number of unserved dynamic customers to realize a minimization of all objectives. Clearly, the second objective is equivalent to the maximization of served dynamic customers.}.

Clearly, the objectives are in conflict and we need to adopt the notion of Pareto-optimality and dominance to describe compromise solutions for the resulting multi-objective optimization problem. For two solutions $x$ and $y$, we denote $x \sdom y$ ($x$ dominates $y$), if $x$ is not worse in any objective and better in at least one objective than $y$. The set of all non-dominated solutions in search space is called \emph{Pareto set}; its image in objective space is called \emph{Pareto front}~\cite{Coello2006}.

As we consider a dynamic problem, anytime a dynamic customer requests service, the Pareto-set would have to be recomputed for the still unvisited mandatory and dynamic customers and a desired solution needs to be selected. As this is usually infeasible in practice,
we discretize time and define $\neras$ intervals of length $\eralength \in \mathbb{R}_{\ge 0}$ called \emph{eras} that partition dynamic requests and subsequent decision making into phases~\cite{RY13}. 

At the onset $t = (j-1)\cdot \Delta$ of each era $j$, new dynamic customer requests may have appeared. Based on this set and the remaining (not yet visited) mandatory customers in $\CM$, we can consider the problem as a static multi-objective optimization problem (MOP) and apply an EMOA to approximate the current Pareto-set. Then, a decision maker (DM) is provided with the compromises and allowed to pick a solution (i.e. a set of $n_v$ tours) which will be realized until the onset of the next era.

Note that in each era $j > 0$ the vehicle has started to realize a tour and possibly has already visited mandatory and/or dynamic customers. Naturally, already realized parts of earlier picked solutions are not reversible anymore.
Thus, decisions made in earlier eras may have significant influence on later solutions. We address this challenge by introducing an automated decision making process as proposed in~\cite{BGMRT2019BiObjective} and evaluating different configurations and decision chains, later on.

\section{A Dynamic Multi-Objective Evolutionary Algorithm}
\label{sec:demoa}
Next we dive into the working principles and algorithmic details of the proposed DEMOA. The algorithm is a natural extension of the DEMOA proposed in \cite{BGMRT2019} for the single-vehicle version of the considered bi-objective problem. The algorithmic steps are outlined in Alg.~\ref{alg:demoa}. The algorithm requires the following parameters: the problem instance comprising of the subsets $\CM$ of mandatory and $\CD$ of dynamic customers. Further parameters control the number and the length of eras ($\neras$ and $\eralength$), the number of vehicles $n_v$ and EA-specific arguments like the population size $\mu$ and parameters controlling for the strength of mutation (details discussed later).
We now describe the DEMOA procedure in general and discuss implementation details (initialization and variation) subsequently.

\begin{algorithm}[htb]
  \caption{DEMOA}
  \label{alg:demoa}
  \begin{algorithmic}[1]
    \Require{\textbf{a)} Customer sets $\CM$, $\CD$, \textbf{b)} nr. of eras $\neras$, \newline{}\textbf{c)} era length $\eralength$, \textbf{d)} nr. of vehicles $\nvehicles$, \textbf{e)} population size $\popsize$, \newline{}\textbf{f)} prob. to swap $\pswap$, \textbf{g)} nr. of swaps $\nswap$}
    \State $t \gets 0$ \Comment{current time}
    \State $\population \gets \emptyset$ \Comment{population (initialized below)}
    \State $\tours \gets$ list of tours \Comment{empty at the beginning of 1st era}
    \For{$i$ $\gets$ 1 to $\neras$} \Comment{era loop} \label{algline:eraloop}
        \State $\dtours \gets$ list of $n_v$ partial tours already driven by vehicles at time $t$ extracted from list $\tours$ \Comment{empty in 1st era}
        \State $\population \gets \Call{initialize}{\popsize, \dtours, t, \population}$ \Comment{see Alg.~\ref{alg:initialize}; pass last population of previous era as template}\label{algline:call_initialize}
        \While{stopping condition not met} \Comment{EMOA loop}
            \State $Q \gets \{\Call{mutate}{x, \dtours, \pswap, \nswap} \,|\, x \in \population\}$ \Comment{Alg.~\ref{alg:mutate}}
            \State $Q \gets \{\Call{localsearch}{x} \, | \, x \in \population\}$
            \State $\population \gets \Call{select}{Q \cup \population}$ \Comment{NSGA-II survival-selection}
        \EndWhile
        \State $\tours \gets \Call{choose}{\population}$ \Comment{DM makes choice $\leadsto$ list of $\nvehicles$ tours}
        \State $t \gets t + \eralength$
    \EndFor
  \end{algorithmic}
\end{algorithm}

\begin{algorithm}[htb]
  \caption{INITIALIZE}
  \label{alg:initialize}
  \begin{algorithmic}[1]
  \Require{\textbf{a)} pop. size $\popsize$, \textbf{b)} initial tours $\dtours$, \textbf{c)} time $t$,\newline{} \textbf{d)} template population $\population$}
  \State $Q \gets \emptyset$
  \For{$j \gets 1$ to $\popsize$}
    \If{$\population$ is empty} \Comment{1st era; no template given}
        \State $x.v_i \gets$ random vehicle from $\{1, \ldots, \nvehicles\}$ for all $i \in \CA$
        \State $x.a_i \gets 1$ for all $i \in \CM$
        \State $x.a_i \gets 0$ for all $i \in \CD$
        \State $x.p \gets$ random permutation of $\CA = \CM \cup \CD$
    \Else \Comment{repair template}
        \State $x \gets \population_j \in \population$ 
        \State $x.a_i \gets 1$ for all $i \in \dtours$
        \State $x.v_i \gets $ vehicle nr. assigned to $i$ in $\dtours$
        \State In $x.p$ move sub-sequence of driven tour in $\dtours$ at the beginning for each vehicle.
        \State $x.a_i \gets 1$ for each $i$ in random subset of $\CDnewavailable$ 
    \EndIf
    \State $Q \gets Q \cup \{x\}$
  \EndFor
  \Return{Q}
  \end{algorithmic}
\end{algorithm}

\subsection{General (D)EMOA}
Initialization steps (Alg.~\ref{alg:demoa}, lines 1-3) consist of declaring a population $P$ and a list $T$ where $T_v$ contains the tour of the corresponding vehicle $v \in \{1, \ldots, \nvehicles\}$, i.e., $T$ stores the solution the decision maker picked at the end of the previous era. Before the first era begins these tours are naturally empty since no planning was conducted at all. Line \ref{algline:eraloop} iterates over the eras. Here, the actual optimization process starts.
The first essential step in each era is -- given the passed time $t$ -- to determine for each vehicle $v \in \{1, \ldots, n_v\}$ the initial tour already realized by vehicle $v$. This information is extracted from the list of tours $\tours$ and stored in the list $\dtours$. Note that again, in the first era the initial tours are empty, as is $T$, since the vehicles are all located at the start depot. Next, the population is initialized in line \ref{algline:call_initialize} and a static EMOA does his job in lines $7-10$. Here, offspring solutions are generated by mutation followed by a sophisticated genetic local search procedure with the aim to reduce the tour lengths of solutions. Finally, following a $(\mu + \lambda)$-strategy the population is updated. Therein, the algorithm relies on the survival selection mechanism of the NSGA-II algorithm~\cite{Deb02}. Once a stopping condition is met, e.g., a maximum number of generations is reached, the era ends and the final solution set is presented to a decision maker who needs to choose exactly one solution~(Alg.~\ref{alg:demoa}, line~11). This choice (the list $\tours$ of $n_v$ tours) determines the (further) order of customers to be visited by the respective service vehicles.

\begin{algorithm}[ht]
    \caption{MUTATE}
    \label{alg:mutate}
    \begin{algorithmic}[1]
    \Require{\textbf{a)} individual $x$, \textbf{b)} initial tours $\dtours$, \textbf{c)} swap prob. $\pswap$, \textbf{d)} $\nswap$}
    \State $\CD_{av} \gets$ dyn. customers available at time $t$ and \underline{not in} $\dtours$
    \State $\CA_{av} \gets$ all customers \underline{av}ailable at time $t$ and \underline{not in} $\dtours$
    \State $\pflipcustomer \gets 1 / |\CD_{av}|$
    \State $\pflipvehicle \gets 1 / |\CA_{av}|$
    \State Flip $x.a_i$ with prob. $\pflipcustomer$ for all $i \in \CD_{av}$
    \State Change vehicle $x.v_i$ with prob. $\pflipvehicle$ for all $i \in \CA_{av}$
    \If{random number in $(0, 1) \leq \pswap$}
        \State Exchange $\nswap$ times each two nodes from $\CA_{av}$ in $x.t$
    \EndIf
    \Return{x}
    \end{algorithmic}
\end{algorithm}

\subsection{Initialization} 
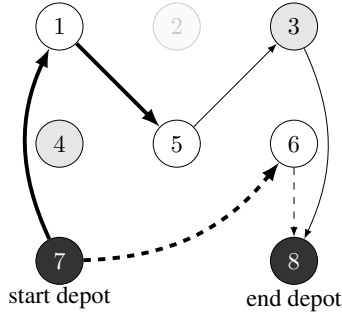
\begin{figure}[htb]
  \centering
  \scalebox{0.92}{
  \begin{tikzpicture}[scale=1,label distance=-6mm]
    \begin{scope}[every node/.style={draw=white}]
    \node (enc) at (0, 0) {
        \begin{tabular}{lcccccc}
        \multicolumn{7}{l}{\textbf{Customer sets}} \\
        \multicolumn{7}{l}{$\CM = \{1,5,6\}, \CD = \{2, 3, 4\}$} \\
        \multicolumn{7}{l}{$\CDavailable = \{3,4\}$} \\
        \multicolumn{7}{l}{} \\
        \multicolumn{7}{l}{\textbf{(Partial) tours}} \\
        \multicolumn{7}{l}{$T_1 = (1,5,3), T_2 = (6)$} \\
        \multicolumn{7}{l}{$\dtours_1 = (1,5), \dtours_2 = (6)$} \\
        \multicolumn{7}{l}{} \\
        \multicolumn{7}{l}{\textbf{Encoding of solution $x$}} \\
        $i$     & 1 & 2 & 3 & 4 & 5 & 6 \\
        \midrule
        $x.v$   & 1 & 2 & 1 & 1 & 1 & 2 \\
        $x.a$   & 1 & 0 & 1 & 0 & 1 & 1 \\
        $x.p$   & 4 & 1 & 6 & 5 & 3 & 2 \\
        \end{tabular}
    };    
    \end{scope}
    \begin{scope}[every node/.style={circle, draw=black, inner sep=4pt}]
    \node[fill=black!80, label={[yshift=-1.2cm]start depot}] (depot1) at (3.2, -1.5) {\textcolor{white}{$7$}}; 
    \node[fill=gray!20, above = 1cm of depot1] (v4) {$4$};
    \node[right = 1cm of v4] (v5) {$5$};
    \node[right = 1cm of v5] (v6) {$6$};
    \node[above = 1cm of v4] (v1) {$1$};
    \node[fill=gray!20, opacity = 0.2, above = 1cm of v5] (v2) {$2$};
    \node[fill=gray!20, above = 1cm of v6] (v3) {$3$};
    \node[fill=black!80, below = 1cm of v6, label={[yshift=-1.2cm]end depot}] (depot2) {\textcolor{white}{$8$}}; 

    \draw (depot1) edge[-latex, ultra thick, bend left=25] (v1);
    \draw (v1) edge[-latex, ultra thick] (v5);
    \draw (v5) edge[-latex] (v3);
    \draw (v3) edge[-latex, bend left=25] (depot2);
    \draw (depot1) edge[-latex, dashed, ultra thick, bend right=25] (v6);
    \draw (v6) edge[-latex, dashed] (depot2);
    \end{scope}
    \end{tikzpicture}
  } 
  \caption{Illustration of the encoding of an individual $x$. Here, customers $i \in \{1, 3, 5, 6\}$ are active ($x.a_i = 1$) while customers $i \in \{2, 4\}$ are inactive ($x.a_i = 0$); customer 4 however asked for service already since $4 \in \CDavailable$. In contrast, customer $2$ did not ask for service so far (illustrated by reduced opacity in the plot). The vehicles already visited a subset of customers (illustrated with thick edges): vehicle one serviced customers 1 and 5 (thus $\dtours_1 = (1, 5)$) while customer 6 was visited by vehicle 2 (thus $\dtours_2 = (6)$).}
  \label{fig:encoding}
\end{figure}

We strongly advice the reader to consult Alg.~\ref{alg:initialize} and in particular Fig.~\ref{fig:encoding} in the course of reading the following explanations for visual support.
Each individual is built of three vectors $x.v, x.a$ and $x.p$ of length $N-2$ each of which stores information on the \underline{v}ehicles assigned, the \underline{a}ctivation status of each customer and a \underline{p}ermutation of all customers. In the first era there are no already visited customers, i.e. both $P$ and $\dtours$ are empty, and the algorithm does not need to take these into account. Hence, in lines 4-7 each individual $x \in \population$ is created from scratch as follows: each customer $i \in (\CM \cup \CD)$ is assigned a vehicle $v \in \{1, \ldots, \nvehicles\}$ uniformly at random. This information is stored in the vector $x.v$. Next, all mandatory customers $i \in \CM$ are activated by setting the value of a binary string $x.a \in \{0, 1\}^{N-2}$ to 1 which means \enquote{active}. In contrast, all dynamic customers $i \in \CD$ are deactivated ($x.a_i = 0$) since they did not ask for service so far. The final step is to store a random permutation of all customers in the permutation vector $x.p$. Note that during fitness function evaluation in order to calculate the individual tour lengths $L_v(x)$ for each vehicle $v$ only the sub-sequence of positions $i \in \{1, ..., N-2\}$ in $x.p$ is considered for which $x.a_i = 1$ and $x.v_i = v$ holds.

Initialization in later eras is different and more complex. Now, the parameter $\population$ passed to Alg.~\ref{alg:initialize} -- the final population of the previous era -- is non-empty and its solutions serve as templates for the new population. We aim to transfer as much information as possible. However, usually the majority of individuals $x \in P$ is in need of repair. This is because as time advances (note that additional $\eralength$ time units have passed) further customers $i \in \CA$ may already have been visited by the vehicle fleet, but it is possible that some of these are inactive in $x$ (i.e., $x.a_i = 0$). Hence, (a) all customers which have already been visited, stored in the list $\dtours$, are activated and assigned to the responsible vehicle (here we use an overloaded element-of relation on lists in the pseudo-codes for convenience) and (b) furthermore their order in the permutation string $x.p$ is repaired. The latter step is achieved by moving the sub-sequence of visited customers before the remaining customers assigned to the corresponding vehicle in the permutation string. This step completes the repair procedure and the resulting individual is guaranteed to be feasible. Subsequent steps involve randomly activating customers that asked for service within the last $\eralength$ time units. 

\subsection{Offspring generation}
The mutation operator (see Alg.~\ref{alg:mutate}) is designed to address all three combinatorial aspects of the underlying problem, i.e., vehicle re-assignment, customer (de)activation and tour permutation. Here, special attention has to be paid to not produce infeasible individuals. Therefore, mutation operates on the subset of customers which have asked for service until now and have not yet been visited. More precisely, each dynamic customer $i \in \CD_{av} = (\CDavailable \setminus \dtours)$ which has not yet been visited is (de)activated with a small probability $\pflipcustomer$ (note that we treat the list $\dtours$ as a set here for convenience). Likewise, each of the customers $i \in \Cavailable = (\Cavailable \setminus \dtours)$ is assigned another vehicle independently with equal probability $\pflipvehicle$. The mutation probabilities $\pflipcustomer$ and $\pflipvehicle$ are set dynamically such that in expectation only one (de)activation or (re)assignment happens; small changes are preferred. Finally, with probability $\pswap$ the permutation vector $x.p$ undergoes $\nswap$ sequential exchange/swap operations (limited to customers which are not fixed so far).
Occasionally, at certain iterations, a local search (LS) procedure is applied to each individual $x \in \population$ (see line~9 in Alg.~\ref{alg:demoa}). The LS takes the vehicle mapping $x.v$ and customer activation $x.a$ as fixed and aims to improve the individual path length by means of the sophisticated solver EAX~\cite{nagata_powerful_2013} for the Traveling-Salesperson-Problem (TSP). To accomplish this goal, given a solution $x \in P$ and a vehicle $v \in \{1, \ldots, \nvehicles\}$ all customers assigned to $v$ in $x$ (appending start and end depot) and their pairwise distances are extracted (nodes 1, 3, 5, 7, 8 for vehicle~1 in Fig.~\ref{fig:encoding}). Next, since the optimization of each vehicle tour is a Hamilton-Path-Problem (HPP) on the assigned customers (no round-trip tour), a sequence of distance matrix transformations is necessary such that the TSP solver EAX can be used to find an approximate solution to the HPP~(see \cite{jonker:transforming} for details). Again, since partial tours might already have been realized, the HPP-optimization starts at the last node visited by the corresponding vehicle (node 5 in Fig.~\ref{fig:encoding} for vehicle~1, since $\dtours_1 = (1,5)$ is fixed already and not subject to changes).

\section{Experimental Methodology}
\label{sec:experiments}

Our benchmark set consists of in total 50 instances with each $n=100$ customers taken from \cite{MGB2015}. There are 10 instances with points spread uniformly at random in the Euclidean plane and each 10 clustered instances with the number of clusters $n_c \in \{2, 3, 5, 10\}$.
The cluster centers are placed by space-filling Latin-Hypercube-Sampling to ensure good spread. Subsequently, $\lfloor n/n_c\rfloor$ nodes are placed around each cluster center assuring cluster segregation with no overlap. We refer the reader to~\cite{MGB2015} for more details on the instance generation process. The proportion of dynamic customers is $50\%$ and $75\%$ for each half of the instance set enabling the study of highly dynamic scenarios.
\begin{table}[htb]
\centering
\caption{DEMOA parameter settings.}
\label{tab:demoa_parameter_settings}
\begin{tabular}{rl}
\toprule
{\bf Parameter} & {\bf Value} \\
\midrule
Nr. of function evaluations per era & $65\,000$ \\
Population size $\popsize$ & 100 \\
Swap probability $\pswap$ & $0.6$ \\
Nr. of swaps $\nswap$ & $10$ \\
Local Search at generations & first, half-time, last \\
\bottomrule
\end{tabular}
\end{table}
\begin{figure}[htb]
    \centering
    \includegraphics[width=0.6\textwidth, trim=0 6pt 0 0, clip]{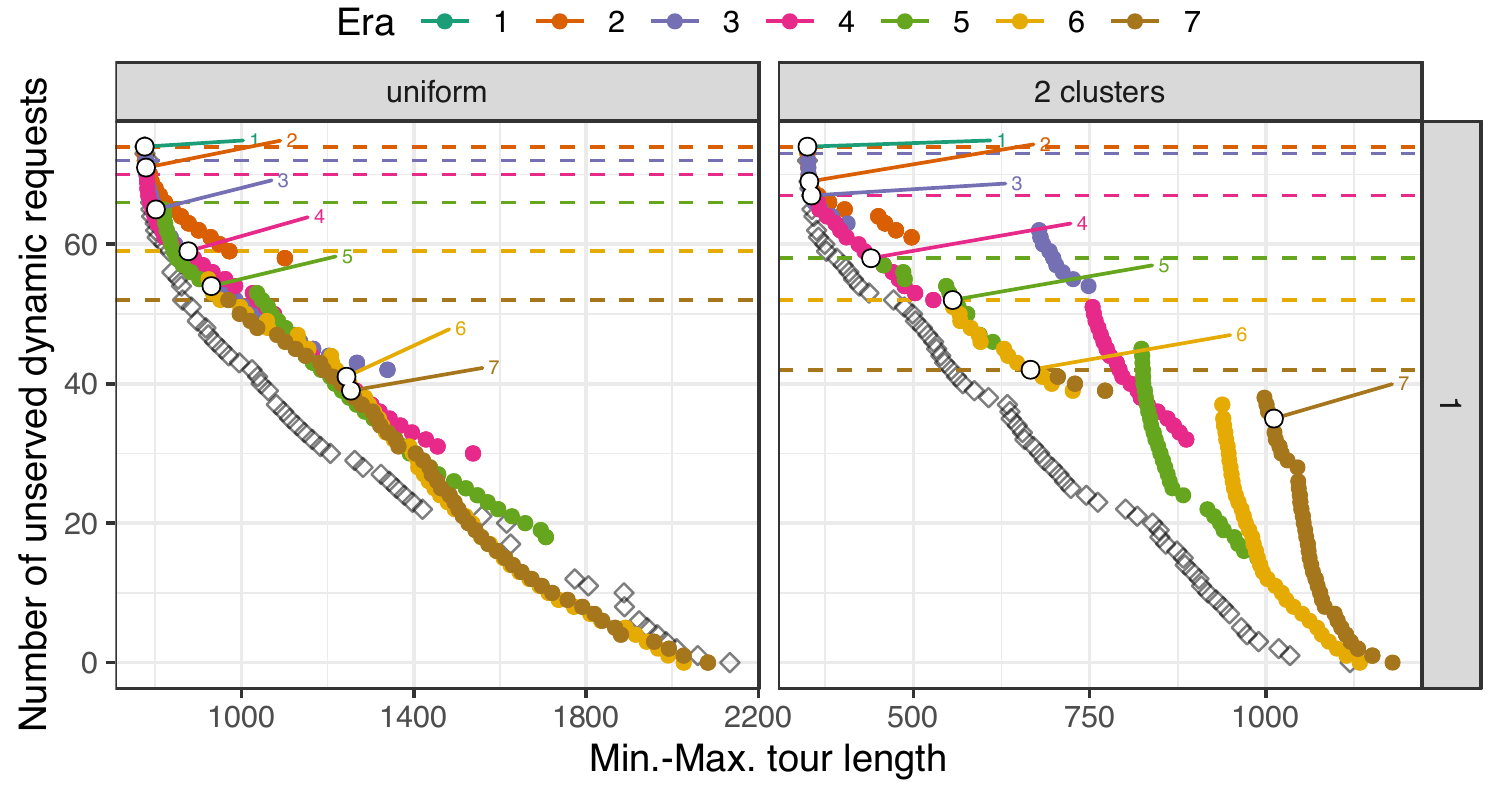}
    \caption{Exemplary Pareto-front approximations for two instances with $75\%$ dynamic requests colored by era and split by problem type (columns) for a single vehicle. Solutions selected by the respective decision maker (\underline{\textbf{0.25-strategy}}) are highlighted and labeled with the era number. Dashed horizontal lines represent the maximum number of dynamic requests which can remain unserved in the respective era.}
    \label{fig:eraplot_examples}
\end{figure}

\subsection{Dynamic aspects and decision making}
We fix $\neras = 7$ eras and set the era-length to $\eralength = \lceil\max_{i \in \CD} r(i) / \neras\rceil$ where $r(i)$ is the request time of customer $i$. $\Delta$ is consistently $\approx 150$ time units across all instances. Naturally, one could start a new epoch once a new customer requests for service. This would result in 50 and 75 eras respectively on our benchmark instances\footnote{Note that the benchmark sets contains instances with $N=100$ customers (\underline{including two depots}) and $\{50\%, 70\%\}$ dynamic customers.}. However, we argue that in a real-world scenario it is more realistic to make decisions after chunks of requests came in and not every single time. For computational experimentation we automate the decision-making process by considering three different decision maker (DM) strategies. To do so, at the end of each era, we sort the final DEMOA population $\population$ in ascending order of the first objective (tour length), i.e. $\population_{(1)} \leq \population_{(2)} \leq \ldots \leq \population_{(\popsize)}$ where the $\leq$-relation is with respect to $f_1$. Note that in the bi-objective space this sorting results in a descending order with respect to the number of unvisited dynamic customers, our second objective $(f_2)$. The automatic DM now picks the solution $\population_{(k)}$ with $k = \lceil d \cdot \popsize \rceil$, $d \in [0, 1]$ where increasing $d$-values correspond to stronger \enquote{customer-greediness}, i.e. higher emphasis on keeping the number of unvisited customers low. In our study we cover $d \in \{0.25, 0.5, 0.75\}$ to account for different levels of greediness and refer to such a policy as a $d$-strategy in the following. Certainly, in real-world scenarios, the DM might change his strategy throughout the day reacting to specific circumstances. However, for a systematic evaluation and to keep our study within feasible ranges, we stick to this subset of decision policies.

\begin{figure}[htb]
    \centering
    \includegraphics[width=0.6\textwidth, trim=0 6pt 0 0, clip]{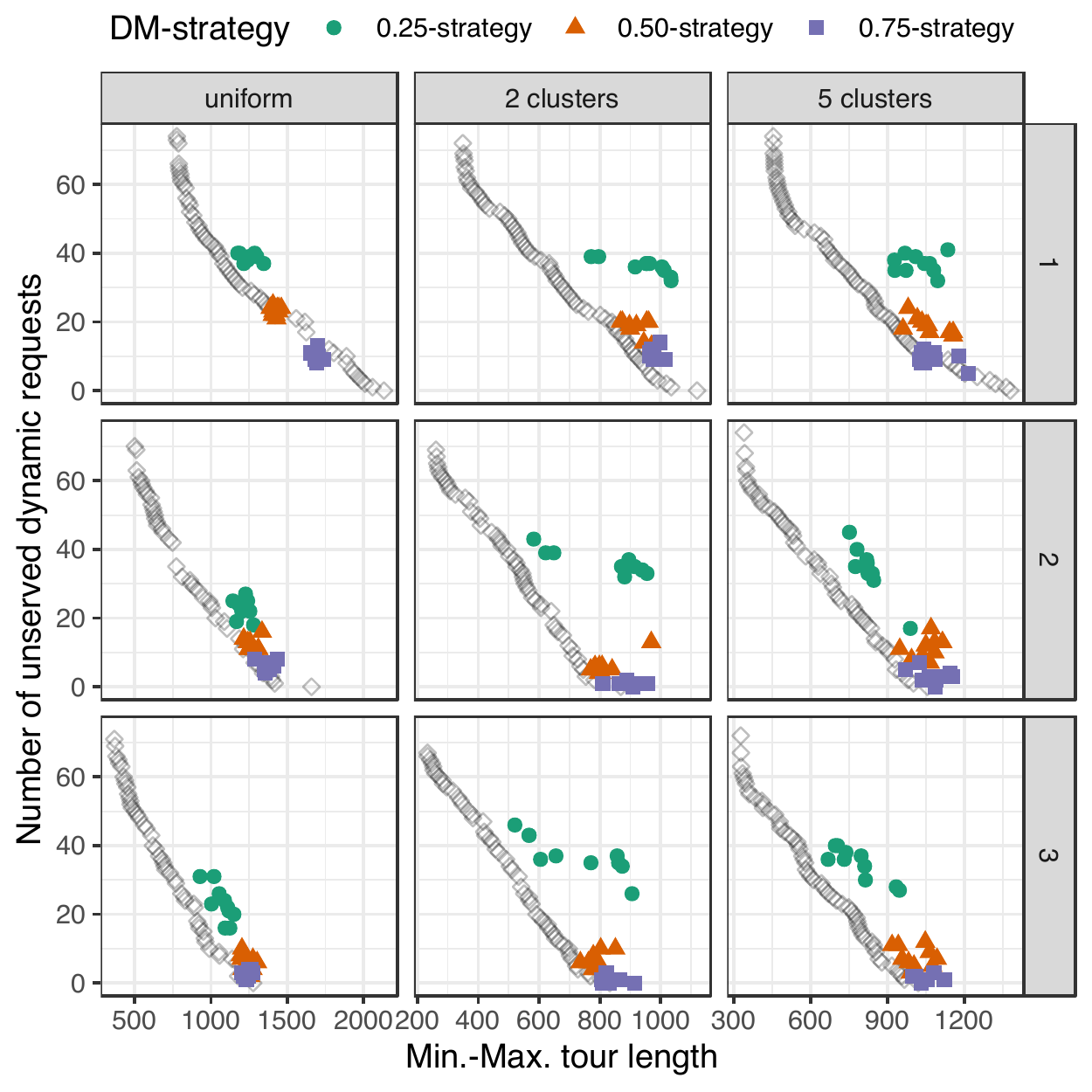}
    \caption{Exemplary visualization of final DM-decisions (i.e., in last era) for three representative instances with $75\%$ dynamic requests colored by decision maker strategy. The data is split by instance (columns) and number of vehicles used (rows).}
    \label{fig:eraplot_by_dmstrategy}
\end{figure}
\subsection{Further parameters}
The further parameter settings of the DEMOA stem from preliminary experimentation and are gathered in Table~\ref{tab:demoa_parameter_settings}.
For each number of vehicles $\nvehicles \in \{1, 2, 3\}$ and each DM strategy $d \in \{0.25, 0.5, 0.75\}$ we run the DEMOA  30 times on each instance for $\neras$ eras following the $d$-strategy. Moreover, for a baseline comparison, we run the clairvoyant EMOA multiple times for each $\nvehicles$ on each instance with a stopping condition of $20\,000\,000$ function evaluations. The clairvoyant EMOA works the same way the DEMOA does. However, it has complete knowledge of the request times of dynamic customers a-priori and treats the problem as a static problem.\footnote{Consider the clairvoyant EMOA as the proposed DEMOA (see Alg.~\ref{alg:demoa}) run for one era with $t=0$ and all customers available from the beginning accounting for request times in the tour length calculations.} This idea was originally introduced in \cite{GMT+2015} for an a-posteriori evaluation of decision making for the single-vehicle variant of the considered problem. We use the Pareto-front approximations of the clairvoyant EMOA as a baseline for performance comparison.
We provide the R implementation of our algorithm in an accompanying repository~\cite{implementation}.

\section{Experimental results}
\label{sec:experimental_results}

In the following, we analyze the proposed dynamic approach (DEMOA) for one to three vehicles with an adapted clairvoyant implementation of the original approach by Bossek et al.~\cite{BGMRT2018}. For a fair comparison, that approach has been extended to deal with multiple vehicles 
and is denoted as EMOA in the following results.

\begin{figure*}[htb]
    \centering
    \includegraphics[width=\textwidth, trim=0 6pt 0 0, clip]{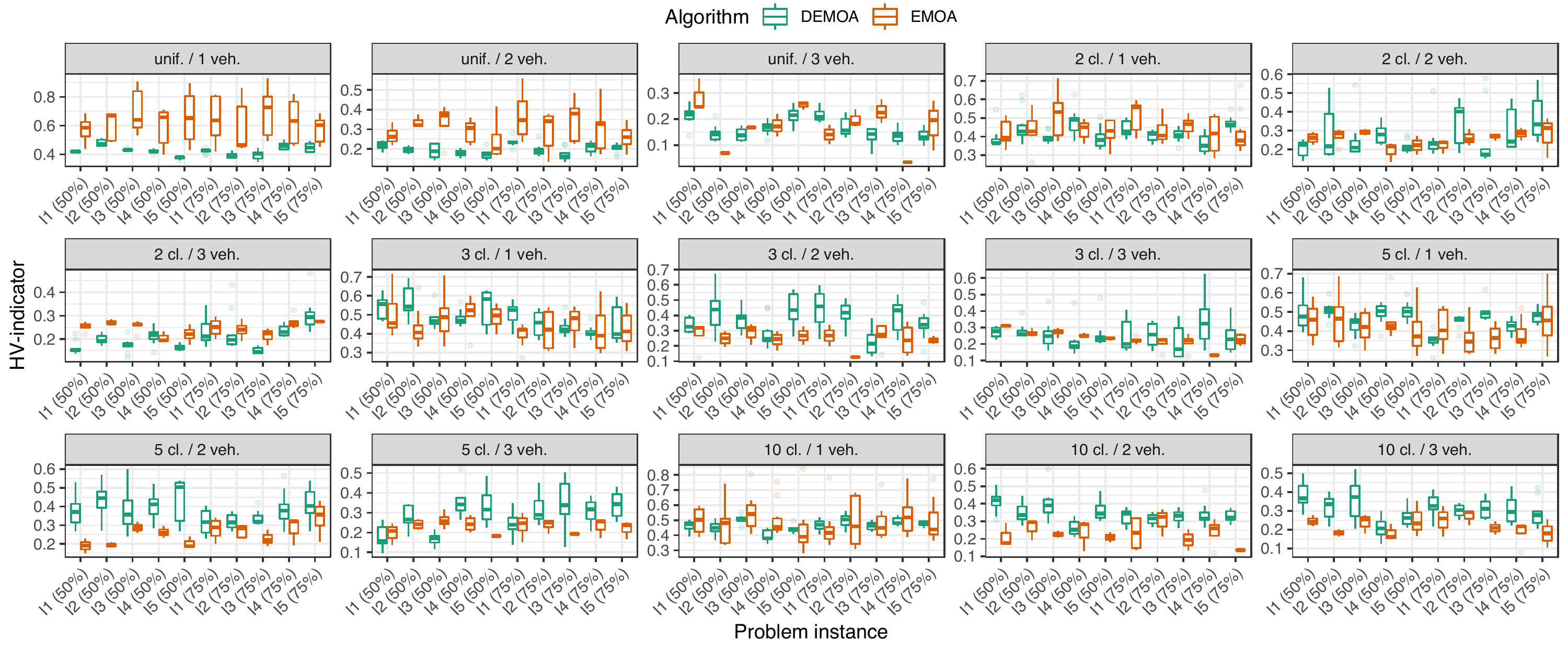}
    \caption{Boxplots of hypervolume-indicator (lower is better) for all instances split by problem type and fleet size. Results are shown for the 0.5-strategy for visual clarity, but the omitted results show the same patterns. Hypervolume values are calculated by (1) instance-wise calculation of the minimal upper bound for the number of unvisited dynamic customers in the last era across all independent runs of the DEMOA, (2) determining the reference point based on the union of all Pareto-front approximations of DEMOA and EMOA chopped at the bound and (3) calculating the HV-value based on this reference set.}
    \label{fig:HV_all}
\end{figure*}

In a first step, we show exemplary results of our dynamic approach to visually introduce the era concept. 
In continuation of the approach of Bossek et al.~\cite{BGMRT2019BiObjective}, we also briefly investigate the influence of decision making to final decision location, when different greediness preferences are considered.
Then, we compare the performance of the EMOA and the proposed DEMOA with respect to three different performance measures. In detail, we investigate the overall performance with respect to problem type and fleet size, the performance with respect to different decision strategies, and the overall performance gain induced by using multiple vehicles.
Finally, we zoom into dedicated solution instances (and their dynamic evolution process) to learn about the behavior of vehicles on clustered and uniform problem instances.

\subsection{Pareto-front approximations and decisions in the dynamic scenario}
\label{sec:exp_dynamic}
The dynamic nature of the problem and online decision making imply that we do not have a single point in time at which the algorithm performance can be evaluated. In the beginning only mandatory customers are available and the tour planning task is equivalent to (multi-vehicle and open) TSP solving. However, as dynamic customer requests appear over time, the initially planned tour(s) must be modified to allow compromises between tour length and number of visited customers. From this point on, a Pareto-set of solutions has to be considered. Following the era concept of decision making, 
at a dedicated point in time, a compromise is chosen for realization by a DM. From that time onward, realization starts and vehicles travel the decided tour. Of course, new dynamic requests appear over time. These are considered at the end of the next era and form a set of new compromises. However, that set of compromises has to consider the already realized partial tour of the vehicles, which cannot be reverted. Consequently, already visited (formerly) dynamic customers reduce the upper bound of unserved customers in compromise sets for future decisions.
The effect of repeated decision making and continuous realization of decisions is exemplarily shown in Figure~\ref{fig:eraplot_examples}. The non-dominated fronts for decisions in all seven eras are shown for a uniform and a clustered instance, respectively. For visual comparison, the clairvoyant EMOA results are also shown. Horizontal dashed lines denote the upper bound of unserved dynamic customers in each stage of decision making. Clearly, in the last era (7, brown points), more customers have been visited by the traveling vehicles than in the first era. Thus, the upper bound has decreased.
\begin{figure*}[ht]
    \centering
    \includegraphics[width=\textwidth, trim=0 6pt 0 0, clip]{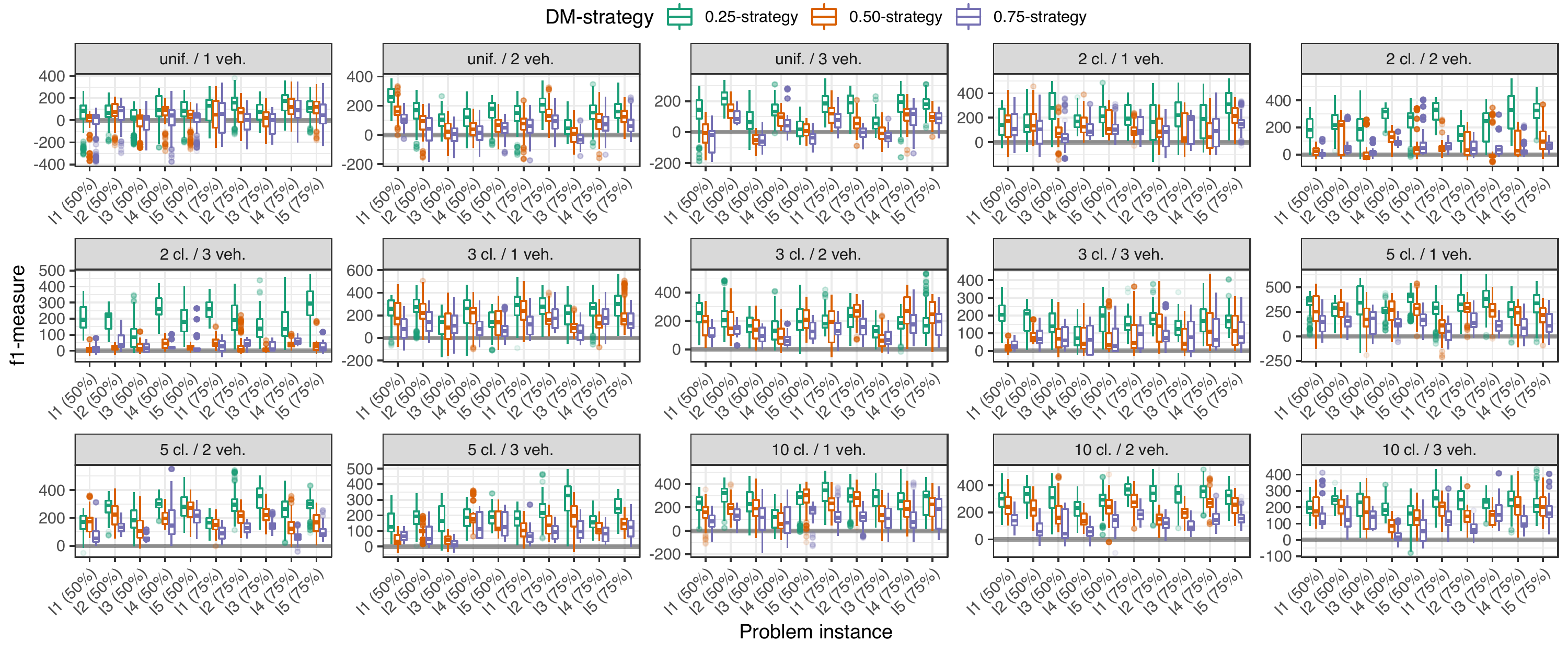}
    \caption{Boxplots of the $f_1$-measure (tour length of DEMOA solution minus tour length of best clairvoyant EMOA solution with the same number of unvisited customers) calculated on basis of all solutions in the last era.}
    \label{fig:f1measure_all}
\end{figure*}
The upper bound and the range of possible decisions is also depending on the decision strategy. A greedy strategy, which aims to reduce the number of unserved customers will favor solutions with many served customers and thus influence realization of longer tours. Less greedy strategies will favor realizations with shorter tours and less visited customers. This behavior was already observed by Bossek et al.~\cite{BGMRT2019BiObjective} for a single vehicle. In Figure~\ref{fig:eraplot_by_dmstrategy}, we confirm an analogous behavior also for multiple vehicles and our algorithmic approach. Therein, visualizations of the results of different strategies (0.25, 0.5, and 0.75 priority of the second objective) and multiple runs for different topologies (uniform and clustered) as well as for different numbers of vehicles are shown. We find, that decision strategies are reflected in the final decision locations. The less greedy strategy produces solutions with more unserved customers than very greedy strategies - independent of the vehicle number.

\subsection{Dynamic and clairvoyant performance and the influence of multiple vehicles}
In order to evaluate the approximation quality of the DEMOA, the dynamic nature of the problem has to be respected. From the decision maker's perspective, the last era (and thus the last non-dominated front) includes all previous decisions and can be compared with the clairvoyant results.
\begin{figure*}[htb]
    \centering
    \includegraphics[width=\textwidth]{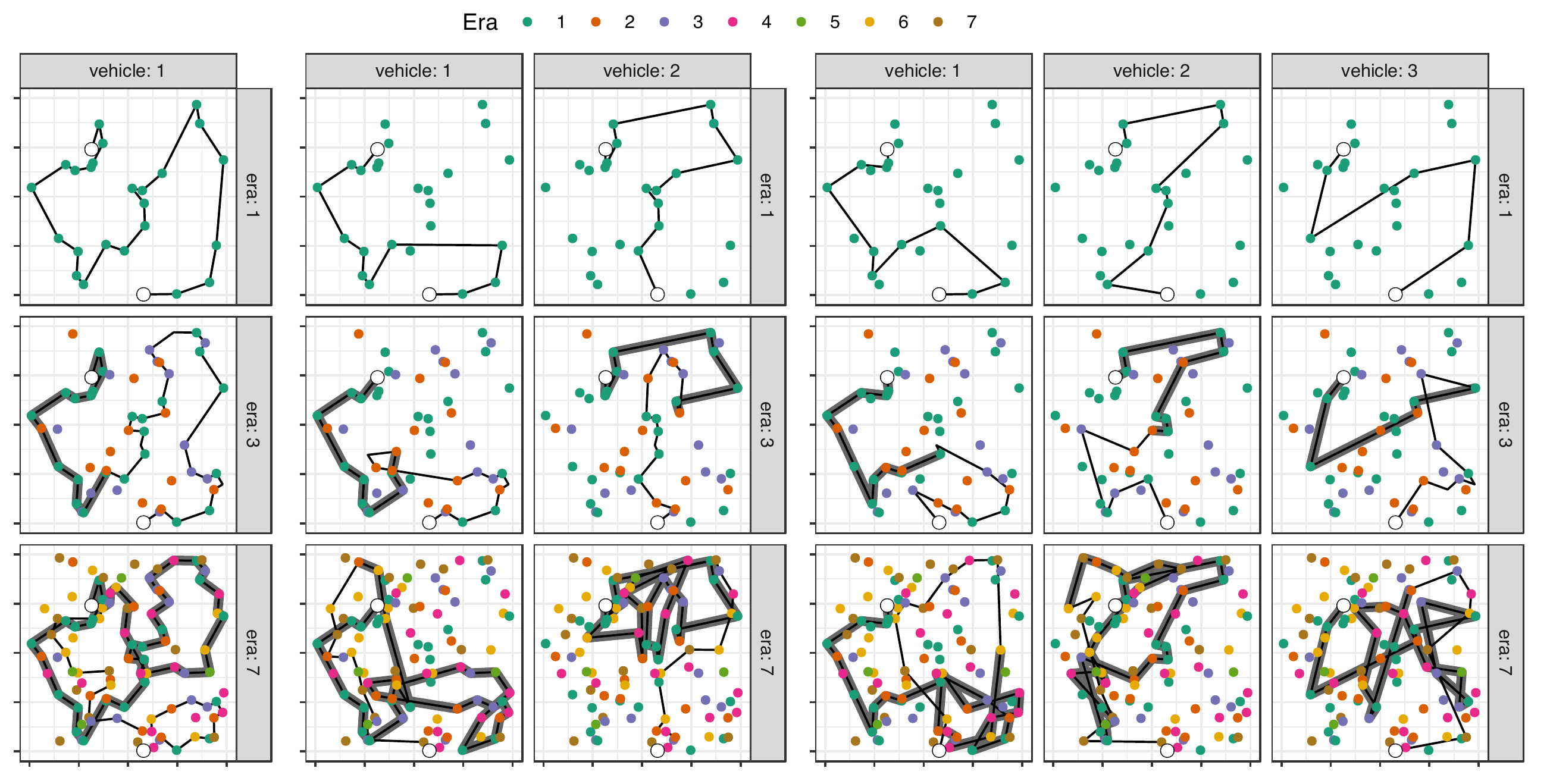}
    \includegraphics[width=\textwidth]{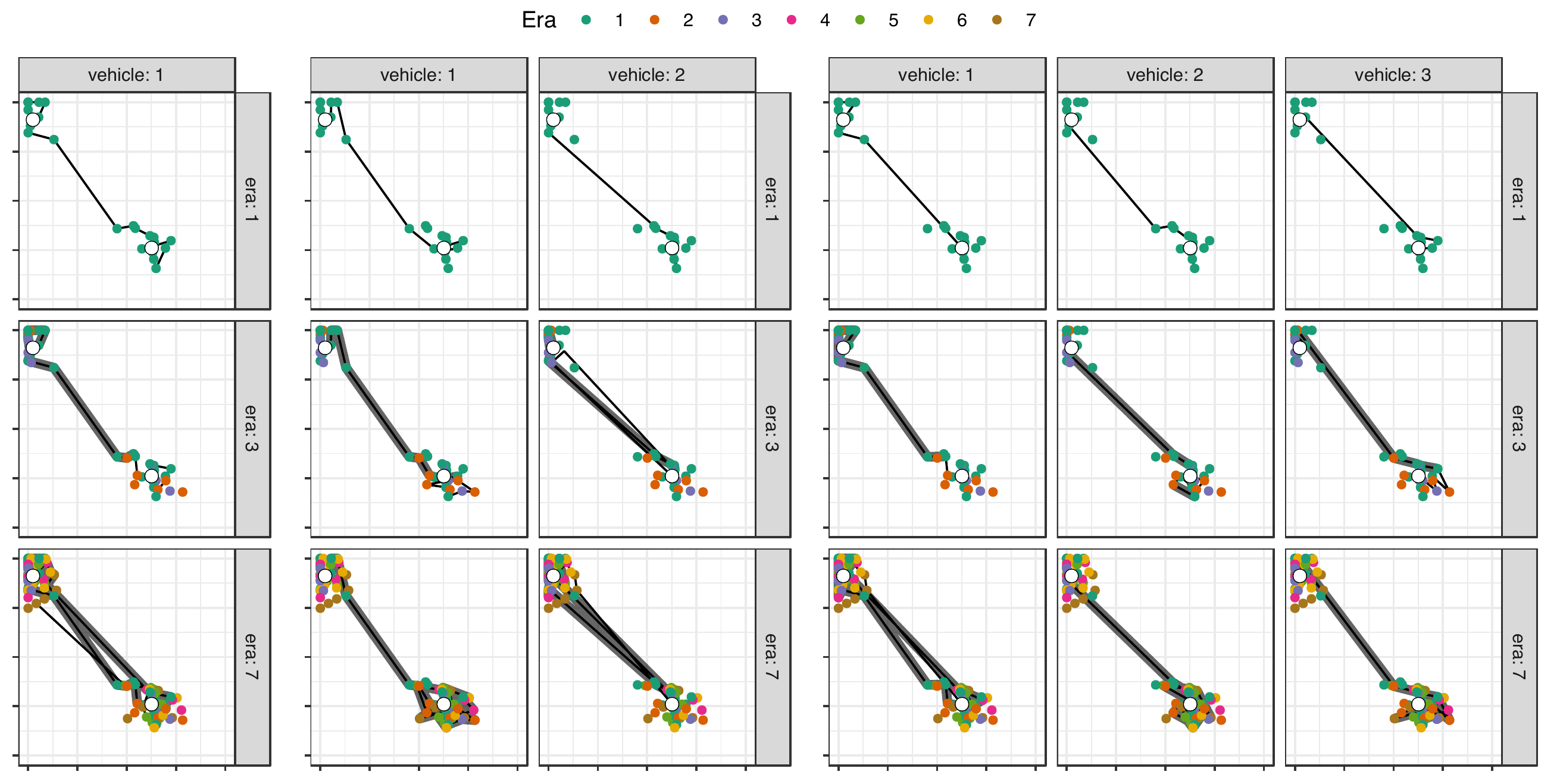}
    \caption{Exemplary tours in eras 1, 3 and 7 (final) with $0.75$-strategy for a uniform instance (top rows) and a clustered instance (bottom rows). For each instance we show the tours the DM picked for a scenario of a single vehicle (left-most column), two vehicles (columns 2 and 3) and three columns (remaining three columns). Bold edges represent the irreversible tour parts which already have been realized by the corresponding vehicle (columns) in the respective era (rows). }
    \label{fig:examplary_tours}
\end{figure*}
\begin{figure}[htb]
    \centering
    \includegraphics[width=0.7\textwidth, trim=0 6pt 0 0, clip]{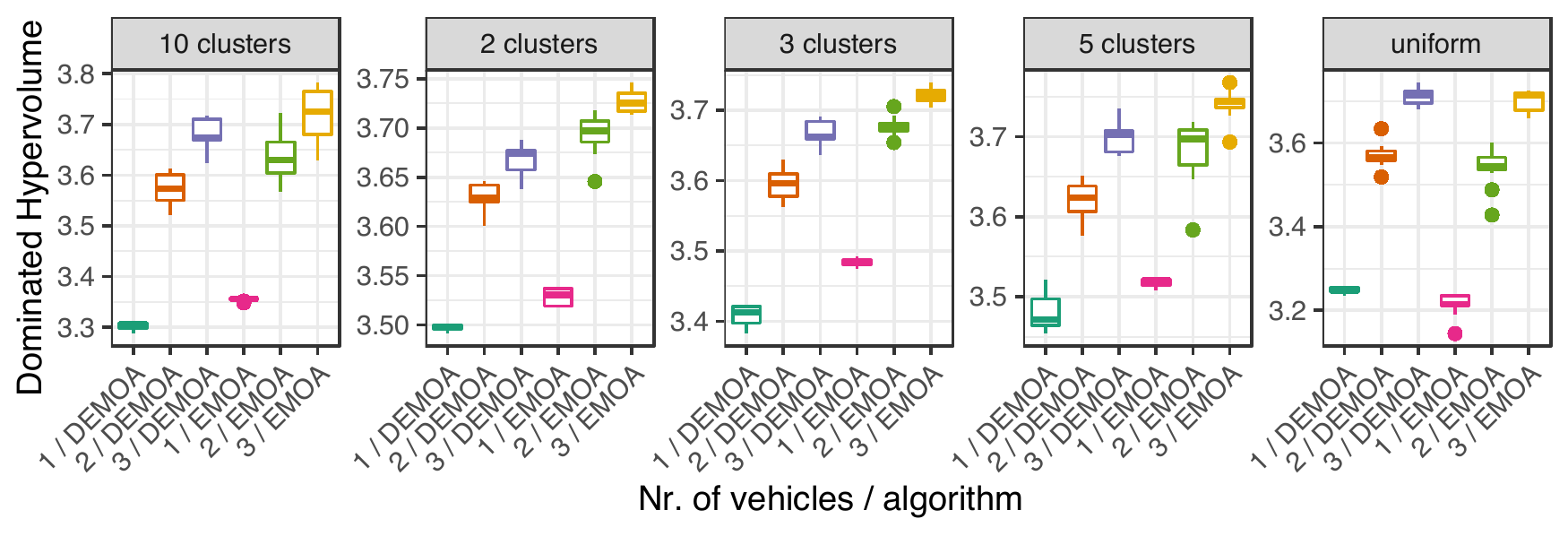}
    \caption{Classical dominated hypervolume distributions (higher is better) of representative instances. The trend is the same for all 50 benchmark instances.}
    \label{fig:hv_vehicle_comparison}
\end{figure}
However, due to the continuously decreasing upper bounds of unserved customers during the optimization process (see~\ref{sec:exp_dynamic}), the EMOA approximation covers a wider range of solutions 
than the  approximated Pareto-front of the final era.
This is considered by our comparison of equivalent ranges of the DEMOA and EMOA results. 

In Figure~\ref{fig:HV_all}, we compare the Hypervolume indicator~\cite{ZDT00} of the DEMOA and EMOA by (1) determining instance-wise the minimal upper bound for the number of unserved customers (objective $f_2$) in the last approximated Pareto-front and for all independent runs of the DEMOA. Then (2), we reduce the solutions of the EMOA to those below the before determined upper bound for $f_2$. From the union of the reduced EMOA results and the DEMOA results, we (3) compute a reference point for Hypervolume computation. Here we compare both algorithms for the medium greedy $0.5$ strategy. We find, that the DEMOA results outperform the EMOA results for uniform problem instances with both $50\%$ and $75\%$ dynamic customers. For clustered instances, performance is in the same range but seemingly depending on the specific instance topology and dynamic customer ratio and service request times. However, we can conclude that the results of the DEMOA for the more realistic dynamic scenario are not necessarily worse than those of the clairvoyant approach. While the clairvoyant EMOA approach knows about the request times of all dynamic customers at $t=0$ and considers all potential customers in compromise generation, the DEMOA optimizes tours of era $i+1$ based on partially realized (and unchangeable) tours from era $i$. This often reduces the size of the tour planning problem significantly and allows to gain very good solutions - some of those even outperform the EMOA solutions in uniform instances.

Another observation from Figure~\ref{fig:HV_all} shows that the aforementioned advantage of problem complexity reduction due to dynamism becomes neglectable, when both algorithms consider multiple vehicles. With increasing number of vehicles, the EMOA tends to outperform the approximation quality of the DEMOA. 

While we focused on a single decision strategy in the previous discussion, Figure~\ref{fig:f1measure_all} investigates  multiple strategies together with tour length properties. Instead of the hypervolume, we now focus on the tour length objective ($f_1$) and measure the distance of solutions to the clairvoyant solution. The idea behind this measure is, that (in the range of the common objective space of DEMOA and EMOA) usually any number of unvisited dynamic customers is covered by the approximated Pareto-front. The convergence quality of a solution set can thus also be expressed by the difference of DEMOA tour length and EMOA tour length. If DEMOA outperforms EMOA, the respective value is negative. The equilibrium of solution quality is denoted by a gray vertical line at $0$ in Figure~\ref{fig:f1measure_all}.  

While we are able to confirm the general observations from above, we now have detailed insights into the effect of decision strategies. As a clear trend, we find better tour lengths per instance topology and dynamism, when greediness in decision making w.r.t. reduction of number of unserved customers ($f_2$) is increased. This holds for the single and multiple vehicle case. The reason for this observation is similar to the argument discussed before. The more customers have to be serviced, the more complex the tour planning problem becomes for the clairvoyant EMOA, while more and more customers are fixed in the dynamic scenario and do not have to be considered for tour planning anymore.

A dedicated view on the effect of using multiple vehicles in the scenario (online and offline) is provided in Figure~\ref{fig:hv_vehicle_comparison}. We find for all investigated cluster configurations and the uniform distribution of customers, that multiple vehicles are advantageous regarding classical hypervolume comparison\footnote{Here the reference point is determined for each problem (independent of the number of vehicles and the strategy). The dominated hypervolume is then calculated for each algorithm, DM-strategy and number of vehicles.}; the results are statistically highly significant with respect to Wilcoxon-Mann-Whitney tests at significance level $\alpha = 0.001$ in $100\%$ of the cases, i.e. two vehicles are significantly better than one and three better than two. At the same time, we observe that the highest gain in solution quality is associated with the step from one to two vehicles. By including a third vehicle, only little is gained. This is probably rooted in the overhead associated with the distances travelled from the start depot to the first customer and from the last customer to the end depot. These distances occur for each vehicle and have to be travelled no matter how short the remaining tour becomes. Thus, this overhead will naturally bound the amount of reasonably applicable vehicles.

\subsection{Exemplary analysis of vehicle tours}
In this paragraph, we briefly investigate representative examples of generated solutions. Figure~\ref{fig:examplary_tours} details the evolution of tours for a single vehicle as well as two, and three vehicles for a uniform and clustered instance in eras one, three, and 7. In the single vehicle case, we can just observe the dynamic adaptation of the planned tour (thin line) towards the realized (and irreversible) tour (bold lines) over time. In the two and three vehicle scenarios, we can nicely observe, how the vehicles automatically partition the customers space. It is obvious, that no vehicle stays idle. Moreover, we find, that each vehicle is assigned similar workload. 
This behavior is partly rooted in the design of our algorithm. As objective $f_1$ minimizes the maximum tour length across all vehicles, selection pressure forces similar workload to each vehicle. Future research has to clarify, whether this desirable feature from a real-world application perspective is always advantageous from the optimization perspective.

\section{Conclusion and Outlook}
\label{sec:conclusion}
In this work, we successfully extended an already high-performing single-vehicle approach to the more realistic dynamic multiple vehicle scenario and we proposed two measures for comparing DEMOA quality to the performance of the related clairvoyant EMOA variant. We find that the algorithmic enhancements ensure a nice distribution of workload (in terms of tour lengths and number of customers served) between the involved vehicles without the necessity for explicitly optimizing for this kind of balance.
At the same time and especially on instances with random uniformly distributed locations, the DEMOA can even outperform the offline EMOA variant with full knowledge of request times. Due to concurrent realization of planned tours, Pareto front approximations of the DEMOA's decision eras naturally concentrate on constantly shrinking problem sizes. These reduced problems can then be solved more effectively than the complete (offline) problem.  
Also, variations of decision makers' preferences and decision chains were investigated. With increasing degree of "greediness", i.e. a stronger focus on minimizing the number of unserved customers, of course overall tour lengths increase, both in the single- as well as in the multiple vehicle scenario. 
The informative and sophisticated visualization approaches presented here can possibly foster the facilitation of decision processes along the DEMOA run and may offer perspectives for future dynamic, tool-based decision support systems.

\bibliographystyle{unsrt}  
\bibliography{bib}  

\end{document}